\documentclass[conference]{IEEEtran}
\IEEEoverridecommandlockouts
\usepackage{cite}
\usepackage{amsmath,amssymb,amsfonts}
\usepackage[ruled,vlined,linesnumbered]{algorithm2e}
\usepackage{graphicx}
\usepackage{textcomp}
\usepackage{float}
\usepackage{caption}
\usepackage{xcolor}
\usepackage{url}
\def\BibTeX{{\rm B\kern-.05em{\sc i\kern-.025em b}\kern-.08em
    T\kern-.1667em\lower.7ex\hbox{E}\kern-.125emX}}
    
    \usepackage{subcaption}

\DeclareCaptionLabelSeparator{custom}{. }
\begin{document}

\title{Line Art Colorization of Fakemon using Generative Adversarial Neural Networks\\
}

\author{\IEEEauthorblockN{Erick Oliveira Rodrigues}
\IEEEauthorblockA{\textit{Department of Academic Informatics (DAINF)} \\
\textit{Universidade Tecnologica Federal do Parana}\\
Pato Branco, Brazil\\
erickrodrigues@utfpr.edu.br}
\and
\IEEEauthorblockN{Esteban Clua}
\IEEEauthorblockA{\textit{Computer Science Department} \\
\textit{Universidade Federal Fluminense}\\
Niteroi, Brazil\\
esteban@ic.uff.br}
\and 
\IEEEauthorblockN{Giovani Bernardes Vitor}
\IEEEauthorblockA{\textit{Institute of Technological Sciences} \\
\textit{Universidade Federal de Itajuba}\\
Itabira, Brazil\\
giovanibernardes@unifei.edu.br}
%\and
%\and
%\and
}

\IEEEoverridecommandlockouts
\IEEEpubid{\makebox[\columnwidth]{978-1-6654-6156-6/22/\$31.00~\copyright2022 IEEE \hfill} \hspace{\columnsep}\makebox[\columnwidth]{ }}

\maketitle

\IEEEpubidadjcol

\begin{abstract}
This work proposes a complete methodology to colorize images of Fakemon, anime-style monster-like creatures. In addition, we propose algorithms to extract the line art from colorized images as well as to extract color hints. Our work is the first in the literature to use automatic color hint extraction, to train the networks specifically with anime-styled creatures and to combine the Pix2Pix and CycleGAN approaches, two different generative adversarial networks that create a single final result. Visual results of the colorizations are feasible but there is still room for improvement.
\end{abstract}


\section{Introduction}

Colorization of images is an important stage in 2D asset production for digital games, animations and digital content production.  Colorization of grey-level images can be traced back to early 2000's \cite{nie05,tai05} where optimization methods \cite{rodrigues17,fractal} were used to convert the grey-level pixels to target colored pixels. Eventually, machine learning started to be used in the same context. Lipowezky \cite{lipowezky06} used classical classification and feature extraction \cite{rodrigues15,rodrigues18}. Later, approaches shifted towards deep learning \cite{huang22}, which includes the usage of Generative Adversarial Networks (GANs), and is now mainstream.

\begin{figure}[H]
\centering
  \begin{subfigure}{0.42\textwidth}
    \includegraphics[width=\linewidth]{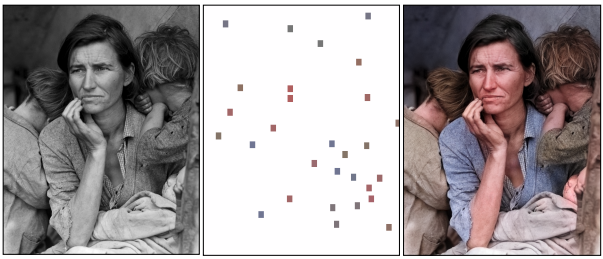}
  \end{subfigure}%
  \caption{Image colorization example by Zhang et al. \cite{zhang17}.} \label{fig:colorex}
\end{figure}

In 2017, Zhang et al. \cite{zhang17} proposed a robust method for image colorization, where one of their results can be seen in Figure \ref{fig:colorex}. One major limitation is that it requires the user to pre-define a ``color palette'', which is also called color hint, as they are associated to the position where the color occur in the image. Their best results cannot be obtained without the color hint.

To this date \cite{xiao19,chen22}, grey-level image colorization is still an open problem. First, we can have different styles of colorization, which leaves room for different approaches. The second issue is image resolution. GANs do not work really well with large resolutions, and the colored images are usually 256x256, eventually 512x512. Large resolutions usually require huge networks that consume lots of memory and require much time to train, while still not producing high quality results.

A third issue is artefact generation. In some cases its possible to spot artefacts such as a repeated colorization pattern that occurs over the entire image. A fourth issue is related to the image domain. Certain types of grey-level images obtain better colorization results. Works coined towards a single type of image tend to excel generic approaches.

It is also possible to find other variations of image colorization in the literature, which includes image colorization from infrared images \cite{kim22,luo22}. However, these are not the primary focus of this work. We are interested in line art colorization for the generation or the acceleration of asset creation for games and entertainment.

Around 2016, we start to get online approaches for line art colorization such as the PaintsChainer \cite{paintschainer}. Line art colorization is a different problem when compared to image colorization. In line art, we just have two main colors as input (black and white - eventually a few shades of gray), as opposed to the image colorization case, where we have rich gray-level information. In this sense, the line art colorization problem requires you to fill in white spaces with varying colors and shades. In line art, the information that can be used is mostly edge, area and location of the pixel (x and y coordinates).

\begin{figure}
\centering
  \begin{subfigure}{0.25\textwidth}
    \includegraphics[width=\linewidth]{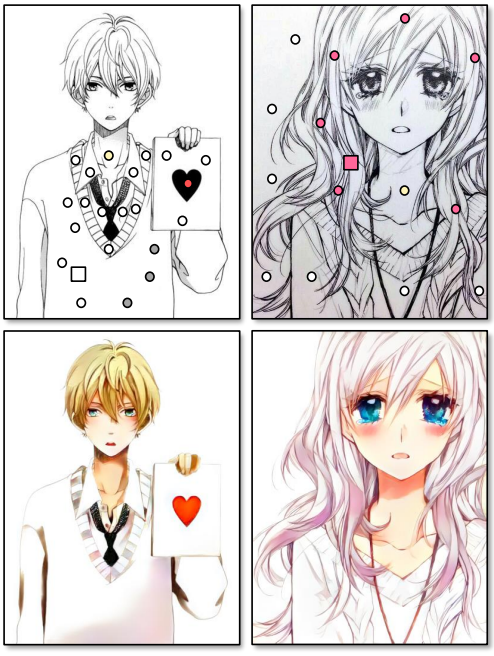}
  \end{subfigure}%
  \caption{Line art colorization example by Zhang et al. \cite{zhang18}.} \label{fig:lineartex}
\end{figure}

Later in 2018, Zhang et al. \cite{zhang18} also proposed an approach for line art colorization that is similar to their previous photo colorization \cite{zhang17}. Figure \ref{fig:lineartex} shows one particular example that includes color hints (squares - before the colorization) and adjustment color hints (circles - after the colorization).

Most line art works in the literature focus on ``humanoid'' characters, which is not directly applicable to creatures/monsters in the kinds of Pokemon and Digimon. In this work, we use the so called ``Fakemons'', which are fan-art monsters inspired by the Pokemon and Digimon franchises. This is the first work to pursue this scenario. We also evaluate the usage of automatic color hints, which is also used for training. Besides, we compare and combine the performance of two different GANs, the Pix2Pix network \cite{isola17} as well as CycleGAN \cite{zhu17}. Combining both networks is also a novel contribution that has never been explored.

The main contributions of this work are: (1) proposal of an automatic and adaptive extraction of line arts, (2) proposal of the first automatic proposal for color hint extraction, (2) first time combination of the responses of the Pix2Pix and CycleGAN and (3) first time line art colorization of monster-like creatures. In what follows, we provide a literature review, proposed methodology, obtained results and conclusion.

\section{Literature Review}

%pix2pix

%cyclegan

Ci et al. \cite{ci18} use a GAN and color hints for line art colorization. Similar to \cite{zhang18}, this work also focus on ``humanoid'' characters. Fang et al. \cite{fang20} also use a GAN and consider different styles of colorization. The obtained results are visually pleasing but are mainly anime faces.

The MANGAN approach \cite{silva19} also uses a GAN, where the authors came up with their own methodology to extract the line art from colorized images in order to train the network. Their line art extraction contains a lot of noise and generated lines are thicker than usual. This extraction can influence other works to under-perform. Hence, the comparison may not be the fairest. Furthermore, the authors also work with colorized ``humanoid'' anime characters. 

Another remarkable factor: the authors apply a Gaussian blur to the color hint, which spreads the colors throughout the image. As a matter of analysis, we also applied a Gaussian blur in our approach but it does not heavily influence the end result of Pix2Pix and CycleGAN. A line or a circular spot is sufficient for Pix2Pix to translate the image appropriately.

\begin{figure}
\centering
  \begin{subfigure}{0.42\textwidth}
    \includegraphics[width=\linewidth]{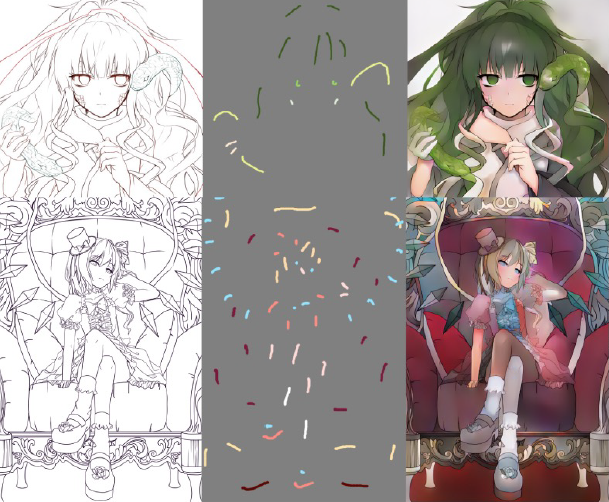}
  \end{subfigure}%
  \caption{Line art colorization result obtained by Hati et al. \cite{hati19}.} \label{fig:lineart_res}
\end{figure}

PaintsTorch \cite{hati19} uses a GAN architecture that is similar to Ci et al.  \cite{ci18}. Their results are pleasing, one of the best results in the literature. However, again, the authors use ``humanoid'' characters and their method uses manually placed color hint. Figure \ref{fig:lineart_res} shows their result. We would like to draw attention to the amount of color hints, which is fairly significant, specially for the image at the bottom.

Serpa et al. \cite{serpa19}, on the other hand, used the Pix2Pix framework to create the shading of 2D sprites in order to accelerate their production. The authors show that they can recreate shading under controlled circumstances. Strange poses increase the difficulties to recreate the shading. This work, however, is not focused on line art colorization. It bears similarities to our approach as it uses Pix2Pix.

Pix2Pix is an image-to-image translation framework based on a GAN. In contrast to Convolutional Neural Networks (CNNs), GANs learn a loss function that classifies whether the output image is real or fake, at the same time that they train a generative model that minimize the loss \cite{isola17}. Pix2Pix uses a ``U-Net'' based architecture for the generator and a ``PatchGAN'' classifier for the discriminator. The generator creates an image and the discriminator is responsible for checking whether the generated image is fake. These two deep neural networks working together compose the GAN architecture. The idea behind Pix2Pix is to provide a general image-to-image framework that works in all sorts of problems and is not application specific.

%In specific, the authors explore the conditional case of GANs. 

The authors \cite{isola17} mention that classical pixel classification treats the output as ``unstructured'' data, as if the information of the class of a certain pixel does not influence on the classification of neighbouring pixels. Conditional GANs (or cGANs), contrariwise, learn a ``structured loss'' function, where this information ``propagates'' over areas of pixels. Pix2Pix is actually a cGAN. Recent works \cite{element} have also explored this approach coupled with classical machine learning (manually coined feature extraction + classifiers), and obtained great results, which means that classifiers other than neural networks can be used and adapted to the ``structured'' idea. This approach is also called ``connectivity''.

Besides Pix2Pix, we also evaluated the CycleGAN framework, also proposed by Isola \cite{zhu17}, author of the Pix2Pix framework. CycleGAN is not the best framework of choice for our problem, as it actually transfers styles of a group of images to another group of images, it does not provide an image-to-image translation. 
CycleGAN is more recommended to cases where the pair of images (line art and colorized ones) are not aligned. However, we combine Pix2Pix and CycleGAN to produce colorized and shaded images.

%os outros trabalhos usam mais de 2000 samples, e usam muitos color hints

\section{Proposed Methodology}

We collected a total of 880 colorized images of Fakemon characters in websites such as DeviantArt, while also including a few examples made by us. We carefully collected pieces of art that use the Creative Commons license. We would like to highlight that we do not hold the copyright for the creatures shown in this work. The collected images were used to train the Pix2Pix and CycleGAN models. We separated a few ``Fakemons'' to be shown as result in this work that were not included in the training dataset. In the end of the work we provide the credits for the used creatures.

The line arts and color hints were automatically extracted from the collected images, which differs from works in the literature. Some works include the color pallet later in the network architecture or treat it as a separate input image. Contrariwise, we actually paint the line art image using circular spots of color hints and do not alter the standard architecture of the network. Therefore, we actually work with a single input image that contains the line art and the color hints.

\subsection{Line art extraction}

First, we use an automatic methodology for the extraction of the line arts, later visiting each one of the characters to verify if the line art was extracted properly. We manually improved the lines in a few cases. However, in real situations, the line art is produced by the artist and this line art extraction step is not necessary. This extraction was required just to create data for training, as it is nearly impossible to find a dataset or an adequate number of pairs of images containing the line art and its colorized version.

To extract the line art, we use an adaptive threshold, shown in Algorithm \ref{alg:adapthres}. We write the tolerance variable according to the number of pixels that compose the character. Larger and more complex characters tend to have thinner lines while small characters tend to have thicker lines. Thus, we increase the tolerance variable if the amount of pixels in the character is low (i.e., pixels that are not background info). Increasing the tolerance variable means that the algorithm will select darker shades of gray for the threshold value. More info on classical image thresholds can be found at \cite{knuckle}.

\begin{algorithm}
\caption{Adaptive threshold algorithm.}\label{alg:adapthres}
\KwData{\small Input image and the $tolerance$ variable, which is a number chosen by the user.}
\KwResult{\small Thresholded binary image containing the line art.}
\small
Convert the input image to gray\;
Construct the histogram of the input image\;
m $\leftarrow$ Find the grey-level that corresponds to the average of the occurrences in the histogram and set to $m$\;
g $\leftarrow$ Order the shades of gray and include in a vector $g$\;
Get the highest shade of gray $v$ in $g$ that satisfies $v * tolerance < m$\;
Apply the classical threshold operation at the value $v$\;
\end{algorithm}

\subsection{Color hint extraction}

After the extraction of the line art, the color hint is estimated. The approach we used is shown in Algorithm \ref{alg:colorest}. This is a k-Medoid clustering algorithm \cite{kms} that skips the transparent pixels and uses an unconventional distance that we adjusted empirically. The used distance is shown in Equation \ref{eq:dist} and r, g and b represent the color layers. $p_1$ and $p_2$ represent the pixels. 

The distance in Equation \ref{eq:dist} does not use the x and y information, and therefore we may end up with clusters of a single color that are much wider than other colors. After that, we apply the k-Medoid algorithm again. For the first time, shown in Algorithm \ref{alg:colorest}, we use $k=35$ to quantize the images in a total of at most 35 colors. This $k=35$ is empirical, but it depends on the objective. Lower values for $k$ (less colors) would reduce the time spent by the artist when placing manual color hints over the line art.

\begin{algorithm}
\caption{\small Automatic color hint algorithm.}\label{alg:colorest}
\KwData{\small  $p_1$ and $p_2$ being the pixels or points, getHue function calculates the usual hue (base color) and getSat calculates the usual saturation (color intensity).}
\KwResult{Image with the color hints.}
\small
Initialize every $k$ cluster, each cluster starts with a single pixel that is selected randomly\;
Iterate over the non-transparent pixels of the image and associate each one of them to one of the $k$ clusters using the distance in Equation \ref{eq:dist}\;
Recalculate the cluster centroids according to the distance in Equation \ref{eq:dist} (the centroid must be an actual pixel that exists - that is not transparent)\;
Return to step 2 and repeat the process until convergence\;
After convergence, iterate the clusters, get the final centroids and draw a circle with the corresponding color in this position\;
\end{algorithm}

\begin{equation}
\small
\label{eq:dist}
\begin{matrix}
d(p_1,p_2) = (getHue(p_1.r*p_1.r,p_1.g*p_1.g,p_1.b*p_1.b)
\\
-getHue(p_2.r*p_2.r,p_2.g*p_2.g,p_2.b*p_2.b))^2
\\
+(1.5*getSat(0.8*p_1.r*p_1.r, 0.8*p_1.g*p_1.g, p_1.b*p_1.b)
\\
-getSat(0.8*p_2.r*p_2.r, 0.8*p_2.g*p_2.g, p_2.b*p_2.b))^2
\end{matrix}
\end{equation}

\begin{figure}
\centering
  \begin{subfigure}{0.24\textwidth}
    \includegraphics[width=\linewidth]{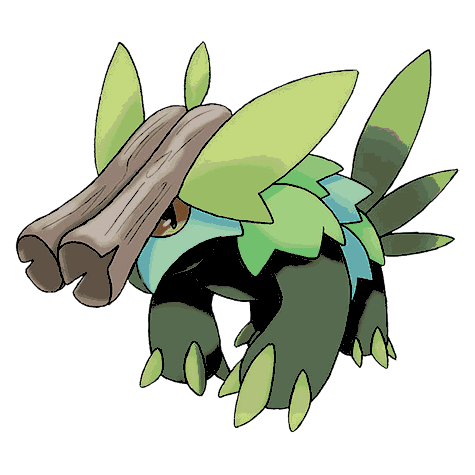}
  \end{subfigure}%
    \begin{subfigure}{0.24\textwidth}
    \includegraphics[width=\linewidth]{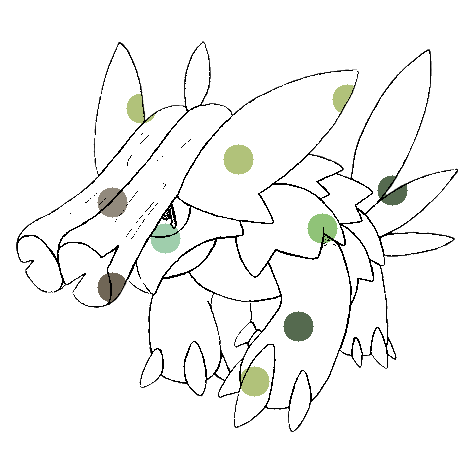}
  \end{subfigure}%

      \begin{subfigure}{0.24\textwidth}
    \includegraphics[width=\linewidth]{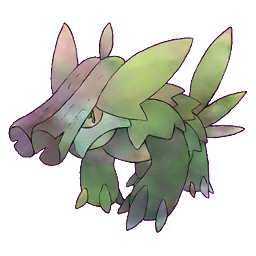}
  \end{subfigure}%
      \begin{subfigure}{0.24\textwidth}
    \includegraphics[width=\linewidth]{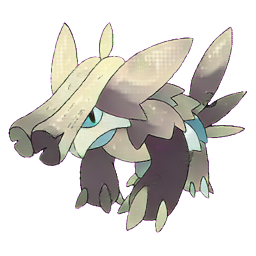}
  \end{subfigure}%
  \label{fig:cluster_fk}
  \caption{The top-left image is the original art by Dragonith (bit.ly/3tQoNLU) quantized to 35 colors. Top-right is the same image after line art and color hint extraction (10 total). Bottom-left is the color result colorized by Pix2Pix and bottom-right by CycleGAN.} 
\end{figure}

Later, for the second time, the k-Medoid algorithm is used with a slight different $k$ and distance so that we can have uniformly spaced different clusters (or color hints) with the same color. The distance used in this second time is the Euclidean distance, with a total of 5 dimensions r, g, b, x and y and equal weights for all of them. For that second time, $k$ is set to 10. Therefore, we automatically generate a total of 10 color hints, where each color hint is represented as a circle of radius 15 in pixels. The result of this processing can be seen in Figure \ref{fig:cluster_fk}. The top-left image in this figure has a total 35 colors (due to the first $k$), and a total of 10 circular color spots (due to the second $k$). The top-right image represents the color hint extraction. The bottom row shows the Pix2Pix and CycleGAN automatic colorizations.

It is possible to increase the number of color hints. However, if we train the algorithm using more color hints, the artist would be required to insert more color hints as well, and we want the colorization to be performed in the easiest way possible. We performed some experiments using more color hints and concluded that the results improve, but we still chose to stick to $k=10$. This trade-off should be analysed carefully.

%The same procedure is repeated for all the 880 images that are used for training. As previously mentioned, we focus on the Pix2Pix framework but we also use CycleGAN despite of the fact that it is not the preferred framework for this task. 
We adjusted the parameters of both Pix2Pix and CycleGAN empirically. The parameters do not influence that much on the final result. However, we increased the ngf and ndf parameters (ngf: number of generator filters in the last conv layer, ndf: number of discriminator filters in the first conv layer) to 150 as they provide better results.

%was extracted using Adobe Photoshop. We first apply 

\section{Results}

\begin{figure*}
\centering
  \begin{subfigure}{0.21\textwidth}
    \includegraphics[width=\linewidth]{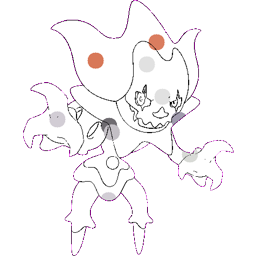}
    \caption{Automatic color hint.}
  \end{subfigure}%
    \begin{subfigure}{0.21\textwidth}
    \includegraphics[width=\linewidth]{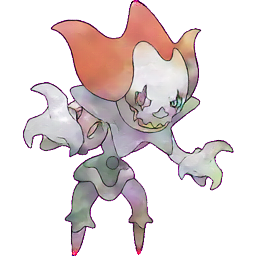}
     \caption{Colorized by the proposed method (Pix2Pix).}
  \end{subfigure}%
    \begin{subfigure}{0.21\textwidth}
    \includegraphics[width=\linewidth]{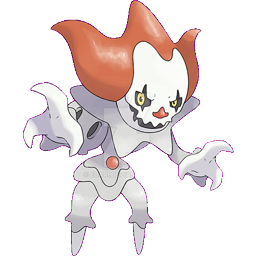}
    \caption{Original art by Edari. https://bit.ly/3y8Vvup}
    \end{subfigure}
      \begin{subfigure}{0.21\textwidth}
    \includegraphics[width=\linewidth]{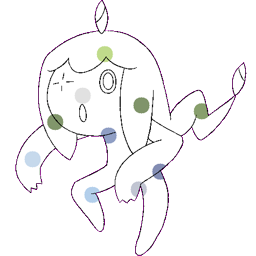}
    \caption{Automatic color hint.}
  \end{subfigure}%

    \begin{subfigure}{0.21\textwidth}
    \includegraphics[width=\linewidth]{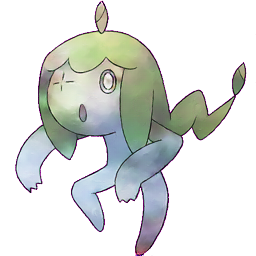}
     \caption{Colorized by the proposed method (Pix2Pix).}
  \end{subfigure}%
    \begin{subfigure}{0.21\textwidth}
    \includegraphics[width=\linewidth]{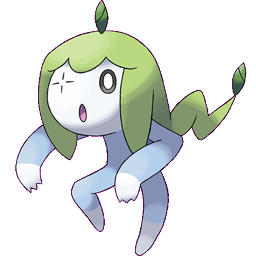}
    \caption{Original art by Bombeetle. https://bit.ly/3QC18sc}
    \end{subfigure}
      \begin{subfigure}{0.21\textwidth}
    \includegraphics[width=\linewidth]{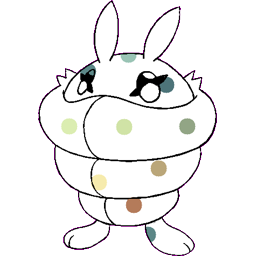}
    \caption{Automatic color hint.}
  \end{subfigure}%
    \begin{subfigure}{0.21\textwidth}
    \includegraphics[width=\linewidth]{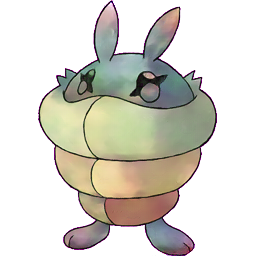}
     \caption{Colorized by the proposed method (Pix2Pix).}
  \end{subfigure}%

    \begin{subfigure}{0.21\textwidth}
    \includegraphics[width=\linewidth]{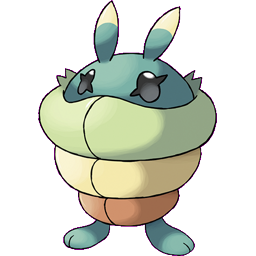}
    \caption{Original art by Bombeetle. https://bit.ly/3QC18sc}
    \end{subfigure}
          \begin{subfigure}{0.21\textwidth}
    \includegraphics[width=\linewidth]{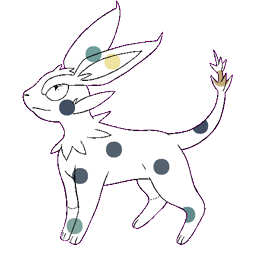}
    \caption{Automatic color hint.}
  \end{subfigure}%
    \begin{subfigure}{0.21\textwidth}
    \includegraphics[width=\linewidth]{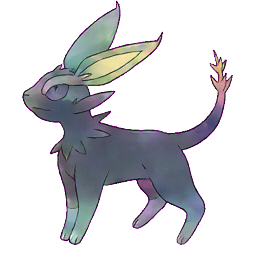}
     \caption{Colorized by the proposed method (Pix2Pix).}
  \end{subfigure}%
     \begin{subfigure}{0.21\textwidth}
    \includegraphics[width=\linewidth]{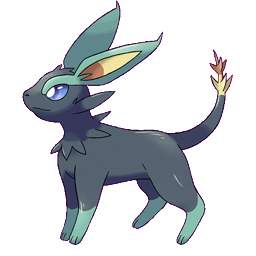}
    \caption{Original art by Bombeetle. https://bit.ly/3xGbOxq}
    \end{subfigure}

  \caption{Results of the colorization obtained with Pix2Pix.}\label{fig:resultpix}
\end{figure*}

Figure \ref{fig:resultpix} shows the colorization obtained with Pix2Pix. These results were obtained directly from the automatic color hint extraction shown in Algorithm \ref{alg:colorest}. We did not use any specific metric to measure the accuracy of the colorization as we adhered to the evaluation performed by other works in the literature, which are based on human visual observation. We can argue that the results are acceptable and that they are suitable for game art and entertainment as is. The style of result kind of reminds watercolor. The boundaries of the images were well colored. No case presented colorizations that exceeded the line art limits.

\begin{figure}
\centering
  \begin{subfigure}{0.26\textwidth}
    \includegraphics[width=\linewidth]{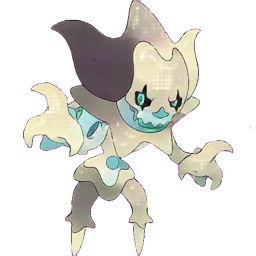}
  \end{subfigure}%
    \begin{subfigure}{0.26\textwidth}
    \includegraphics[width=\linewidth]{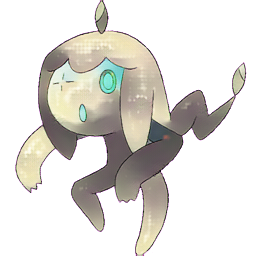}
  \end{subfigure}%

      \begin{subfigure}{0.26\textwidth}
    \includegraphics[width=\linewidth]{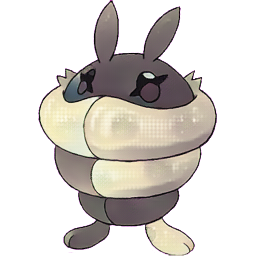}
  \end{subfigure}%
      \begin{subfigure}{0.26\textwidth}
    \includegraphics[width=\linewidth]{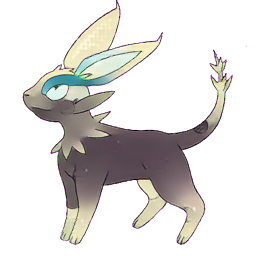}
  \end{subfigure}%
  \caption{CycleGAN result for the same input parameters as Figure \ref{fig:resultpix}.} \label{fig:cycleres}
\end{figure}

Figure \ref{fig:cycleres} shows some results obtained with CycleGAN. We are not sure why the colors did not vary that much, as other the colors varied in other experiments with CycleGAN. One clear difference is that it is capable of grasping the darker and lighter shades better than Pix2Pix, even providing a type of ``dither'' where it should be lighter such as in the heads of the two first Fakemon in Figure \ref{fig:cycleres}. Pix2Pix, on the other hand, works better with the colors.

\begin{figure}
\centering
  \begin{subfigure}{0.26\textwidth}
    \includegraphics[width=\linewidth]{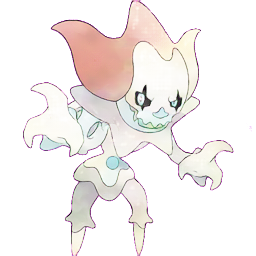}
  \end{subfigure}%
    \begin{subfigure}{0.26\textwidth}
    \includegraphics[width=\linewidth]{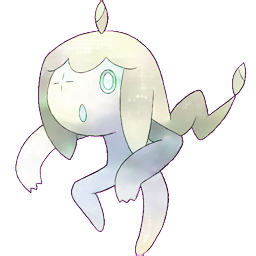}
  \end{subfigure}%

      \begin{subfigure}{0.26\textwidth}
    \includegraphics[width=\linewidth]{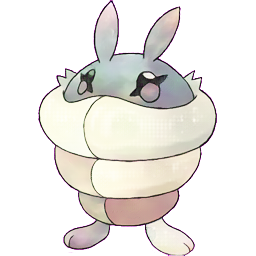}
  \end{subfigure}%
      \begin{subfigure}{0.26\textwidth}
    \includegraphics[width=\linewidth]{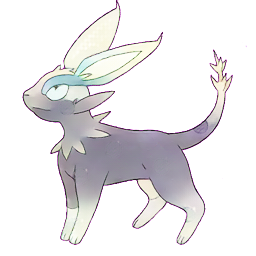}
  \end{subfigure}%
  \caption{Combination of the Pix2Pix and CycleGAN responses (one divided by the other - blend mode in Photoshop). This approach is purely automatic, line art and color hints (10 total, $k=10$) were extracted automatically.} \label{fig:comb}
\end{figure}

Considering that both approaches have interesting aspects (one the color and the other the shading), we tried to combine them. To our surprise, the combination yielded interesting results. If we divide (blend mode) the Pix2Pix result by the CycleGAN result, we get the response shown in Figure \ref{fig:comb}, with soft colors and shading.

%We can improve the results by manually inserting the color hints as other works in the literature. When we manually adjust color hints, we get the results shown in.

\begin{figure}
\centering
  %\begin{subfigure}{0.24\textwidth}
 %   \includegraphics[width=\linewidth]{figures/various/6_fake_B.png}
 % \end{subfigure}%
 %   \begin{subfigure}{0.24\textwidth}
  %  \includegraphics[width=\linewidth]{figures/various/19_fake_B.png}
 % \end{subfigure}%

      \begin{subfigure}{0.24\textwidth}
    \includegraphics[width=\linewidth]{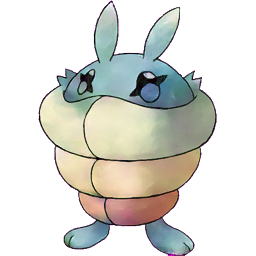}
  \end{subfigure}%
      \begin{subfigure}{0.24\textwidth}
    \includegraphics[width=\linewidth]{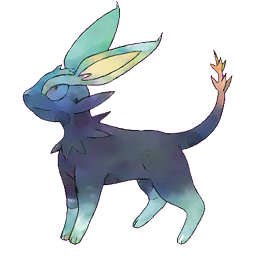}
  \end{subfigure}%
  \caption{Pix2Pix result with manually positioned 30 color hints. Results did not improve substantially.} \label{fig:several}
\end{figure}

As a final experiment, we manually included several color hints and analysed how the Pix2Pix perform. The result can be seen in Figure \ref{fig:several}. Results do improve in some areas of the image. Overall, the improvement is not substantial.

All the experiments in this work were run using an Intel i7-10700 CPU and a Nvidia RTX 3060 with 12GB of memory. The Pix2Pix trained for 2 days, past epoch 1000, while the CycleGAN trained for 4 days straight and reached epoch 300. We experimented with previous epochs to double check for overfitting, but the results did not seem to improve. The parameters were mostly the standard ones, we increased the ngf and ndf parameters to the maximum of memory the GPU would support (150 for Pix2Pix and 128 for CycleGAN).

\section{Conclusion}

This work performs experiments towards the automatic colorization of Fakemons, monster-like characters. We collected a total of 880 images that fit in the ``Fakemon'' category for training. Contributions of this work include the algorithm for the extraction of the line art as well as the automatic extraction and generation of the color hints. Besides, a major contribution is colorizing anime-styled creatures, which is the first occasion in the literature. We are also the first to experiment with a small amount of color hints and to combine Pix2Pix and CycleGAN in the same result.

The first major conclusion is that there is still a lot of room for improvement in automatic colorization. Even with a fairly large dataset we still had to combine the approaches to obtain arguably adequate results (Figure \ref{fig:comb}). The results can still be improved to look more like the original pieces. Furthermore, we also have a limitation with respect to the resolution, the images were 256x256.

However, we also conclude that the results are feasible, they remind the watercolor style and can be used as is in games and entertainment. One particular advantage is that all the generated colorization appear to be colorized with the same art style, irrespective of the author who created it. These approaches can be used, for instance, to ``standardize'' pieces of art from different authors.

We used frameworks (Pix2Pix and CycleGAN) that are generic for this type of task (image to image translation and style transfer), coined to work with all sorts of problems. Future works may also include the creation of frameworks and approaches coined specifically for this task, aiming to improve the results and similarity to the original piece.

Although the colorization of Fakemon and usual anime art may sound like the same thing, they are actually very different approaches. Color palettes are different, line art is different, the background information is different, etc. This justifies the creation of a framework for this type of task as future work, as well as the use of specific images for training.

%As a final remark, 

%Therefore, the literature required an approach of this kind to check how the algorithms would perform.

\section*{Acknowledgment}

All the art shown in this work was extracted from DeviantArt.com. We carefully selected arts that conform to the Creative Commons license, which can be found at their website. In this manuscript, we include pieces from the artists called Dragonith (https://www.deviantart.com/dragonith), Edari (https://www.deviantart.com/edari) and Bombeetle (https://www.deviantart.com/bombeetle), which can be found at their profile page.

%\section*{References}

\bibliographystyle{unsrt}
\bibliography{mybibfile}

\end{document}